\documentclass[aip,cha,amsmath,amssymb,reprint,]{revtex4-1} 
\pdfoutput=1

\usepackage{graphicx} 
\usepackage{dcolumn} 
\usepackage{bm} 
\usepackage[hidelinks,breaklinks]{hyperref} 
\usepackage[capitalize,nameinlink]{cleveref} 
\usepackage{enumitem} 

\newcommand{\hlchangeI}[1]{#1}
\newcommand{\hlchangeII}[1]{#1}
\newcommand{\hlchangeIr}[1]{#1}
\newcommand{\hlchangeIIr}[1]{#1}

\newcolumntype{d}{D{.}{.}{-1}}

\usepackage[utf8]{inputenc}
\usepackage[T1]{fontenc}
\usepackage{mathptmx}

\begin{document}


\title{Forecasting Chaotic Systems with Very Low Connectivity Reservoir Computers}

\author{Aaron Griffith}
\affiliation{Department of Physics, The Ohio State University, 191 W.\ Woodruff Ave., Columbus, Ohio 43210, USA}

\author{Andrew Pomerance}
\affiliation{Potomac Research LLC, 801 N.\ Pitt St.\ \# 117, Alexandria, Virginia 22314, USA}

\author{Daniel J. Gauthier}
\affiliation{Department of Physics, The Ohio State University, 191 W.\ Woodruff Ave., Columbus, Ohio 43210, USA}

\date{\today}

\makeatletter
\hypersetup{pdftitle={\@title},pdfauthor={A.~Griffith, A.~Pomerance, D.~J.~Gauthier}}
\makeatother

\begin{abstract}
We explore the hyperparameter space of reservoir computers used for
forecasting of the chaotic Lorenz '63 attractor with Bayesian
optimization. We use a new measure of reservoir performance, designed
to emphasize learning the global climate of the forecasted system
rather than short-term prediction. We find that optimizing over this
measure more quickly excludes reservoirs that fail to reproduce the
climate. The results of optimization are surprising: the optimized
parameters often specify a reservoir network with very low
connectivity. Inspired by this observation, we explore reservoir
designs with even simpler structure, and find well-performing
reservoirs that have zero spectral radius and no recurrence.  These
simple reservoirs provide counterexamples to widely used heuristics in
the field, and may be useful for hardware implementations of reservoir
computers.
\end{abstract}

\maketitle

\begin{quotation}
Reservoir computers have seen wide use in forecasting physical
systems, inferring unmeasured values in systems, and
classification. The construction of a reservoir computer is often
reduced to a handful of tunable parameters. Choosing the best
parameters for the job at hand is a difficult task. We explored this
parameter space on the forecasting task with Bayesian optimization
using a new measure for reservoir performance that emphasizes global
climate reproduction and avoids known problems with the usual measure.
We find that even reservoir computers with a very simple construction
still perform well at the task of system forecasting. These simple
constructions break \hlchangeII{common} rules for reservoir construction and
may prove easier to implement in hardware than their more complex
variants while still performing as well.
\end{quotation}

\section{Introduction}

A reservoir computer (RC) is a machine learning tool that has been used
successfully for chaotic system forecasting\cite{jaeger1978} and
hidden-variable observation.\cite{lu2017} The RC uses an internal or hidden artificial neural network known as a reservoir, which is a dynamic system that reacts over time to changes in its inputs. Since the RC is a dynamical system with a characteristic time scale, it is a
good fit for solving problems where time and history are critical.

More recently, RCs were used to learn the \emph{climate} of a chaotic
system;\cite{pathak2017,haluszczynski2019} that is, it learns the
long-term features of the system, such as the system's
attractor. Reservoir computers have also been realized physically as
networks of autonomous logic on an FPGA\cite{canaday2018} or as
optical feedback systems,\cite{antonik2016} both of which can perform
chaotic system forecasting at a very high rate.

A common issue that must be addressed in all of these implementations
is designing the internal reservoir. Commonly, the reservoir is
created as a network of interacting nodes with a random topology. Many
types of topologies have been investigated, from
Erd{\"{o}}s-R{\'{e}}nyi networks and small-world
networks\cite{haluszczynski2019} to simpler cycle and line
networks.\cite{rodan2011} Optimizing the RC performance for a specific
task is accomplished by adjusting some large-scale network properties,
known as \emph{hyperparameters}, while constraining others.

Choosing the correct hyperparameters is a difficult problem because the hyperparameter space can be large. There are a handful of known results for some
parameters, such as setting the \emph{spectral radius} $\rho_r$ of the network near to unity and the need for recurrent network connections,\cite{jaeger2001,lukosevicius2012} but the applicability of these results is narrow. In the absence of guiding rules, choosing the hyperparameters is done with costly optimization methods, such as grid search,\cite{rodan2011} or methods that only work on continuous parameters, such
as gradient descent.\cite{jaeger2007}

The hyperparameter optimization problem has also been solved with
Bayesian methods,\cite{yperman2016,maat2018} which are well suited to optimize over either discrete or continuous functions that are computationally intensive and potentially noisy. Hyperparameters
optimized in this way can be surprising: the Bayesian algorithm can
find an optimum set of parameters that defy \hlchangeII{common} heuristics for
choosing reservoir parameters, as we demonstrate below.

We use this Bayesian approach for optimizing RC hyperparameters for the task of replicating the climate of the chaotic Lorenz '63
attractor,\cite{lorenz1963}
\hlchangeI{the R{\"{o}}ssler attractor,\cite{rossler1976} and a chaotic double-scroll circuit.\cite{gauthier1996}}
We introduce a new
measure of reservoir performance designed to emphasize global climate
reproduction as opposed to focusing only on short-term forecasting.  During
optimization, we find that the optimizer often settled on
hyperparameters that describe a reservoir network with extremely low
connectivity, but which function as well as networks with higher
connectivity. Inspired by this observation, we investigate even simpler
reservoir topologies. We discover reservoirs that successfully
replicate the climate despite having no recurrent connections and 
$\rho_r=0$. Such simple network topologies may be easier to synthesize in physical RC realizations, where internal connections and recurrence
have a hardware cost.

The rest of this paper is structured as follows: in
\cref{sec:reservoir}, we describe our RC construction at a high
level. We describe the Lorenz '63 system,
\hlchangeI{the R{\"{o}}ssler system, and a double-scroll chaotic
  circuit in \cref{sec:systems}, which we use as examples for the
  forecasting task.}
In \cref{sec:construct-train},
we detail the specifics of how our reservoir networks are constructed,
introduce the five hyperparameters we consider, and describe the five
network topologies we investigate. We also discuss how to choose these
hyperparameters with Bayesian optimization, and how to train the
resulting RC.\@ \Cref{sec:forecasting} explains our process for
evaluating the forecasting ability of these RCs. We discuss the
short-term forecasting performance measure and its pitfalls and
introduce our modified measure of performance. \Cref{sec:results}
describes the results of our investigation, and finally
\cref{sec:conclusion} concludes with some ideas for applying these
results in future research.

\section{Reservoir Computers}%
\label{sec:reservoir}  

At a high level, an RC is a method to transform
one time-varying signal (the input to the RC) into
another time-varying signal (the output of the RC),
using the dynamics of an internal system called the \emph{reservoir}.

\begin{figure}
  \includegraphics{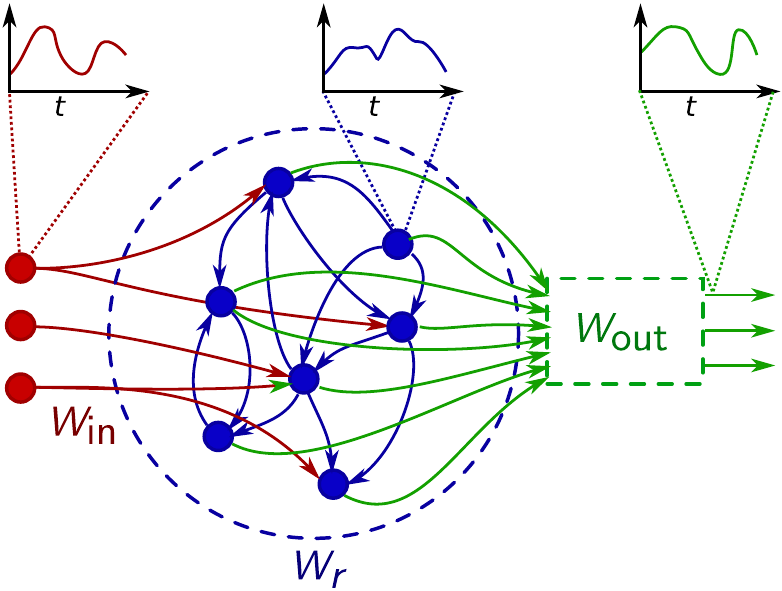}
  \caption{High-level view of a reservoir computer. Each node may have
    three kinds of connections: connections to other nodes in the
    network ($W_r$, blue), connections to the overall input
    ($W_\text{in}$, red), or connections to the output
    ($W_\text{out}$, green). Note that the internal connections may
    contain cycles.  When the RC is used to perform forecasting,
    the output on the right side is connected to the input on the left
    side, allowing the RC to run autonomously with no external
    input.}%
  \label{fig:overview}
\end{figure}

We use an RC construct known as an \emph{echo state network},\cite{jaeger2001} which  uses a network of nodes as the internal reservoir. Every node has inputs, drawn from
other nodes in the reservoir or from the input to the RC, and every input has an associated weight. Each node
also has an output, described by a differential equation. The output
of each node in the network is fed into the \emph{output layer} of the
RC, which performs a linear operation of the node
values to produce the output of the RC as a whole. This
construction is summarized in \cref{fig:overview}.  

\subsection{Reservoir}

The dynamics of the reservoir are described by
\begin{equation}
  \label{eq:reservoir}
  \dot{\mathbf{r}}(t) = -\gamma\,\mathbf{r}(t) + \gamma\,\tanh \left( W_r \mathbf{r}(t) + W_\text{in} \mathbf{u}(t) \right),
\end{equation}
where each dimension of the vector $\mathbf{r}$ represents a single
node in the network.  Here, the function $\tanh(\ldots)$
operates component-wise over vectors: $\tanh{(\mathbf{x})}_i =
\tanh(x_i)$.

For our study, we fix the dimension of the reservoir vector $\mathbf{r}$
at $N = 100$ nodes, and the dimension $d$ of the input signal
$\mathbf{u}(t)$ is $d = 3$. Therefore, $W_r$ is an $N \times N$ matrix
encoding connections between nodes in the network, and $W_\text{in}$
is an $N \times d$ matrix encoding connections between the 
reservoir input $\mathbf{u}(t)$ and the nodes within the
reservoir. The parameter $\gamma$ defines a natural rate (inverse time scale)
of the reservoir dynamics.  The RC performance depends on the specific choice of
$\gamma$, $W_r$, and $W_\text{in}$. This choice is discussed further
in \cref{subsec:construct}.

\subsection{Output Layer}

The output layer consists of a linear transformation of a function of node values
\begin{equation}
  \mathbf{y}(t) = W_\text{out} \mathbf{\tilde{r}}(t),%
  \label{eq:output}
\end{equation}
where $\mathbf{\tilde{r}}(t) = \mathbf{f_\text{out}}\left(\mathbf{r}(t)\right)$. The function
$\mathbf{f_\text{out}}$ is chosen ahead of time to break any unwanted
symmetries in the reservoir system. If no such symmetries exist,
$\mathbf{\tilde{r}}(t) = \mathbf{r}(t)$ suffices. $W_\text{out}$ is
chosen by \emph{supervised training} of the RC.\@ First, the reservoir
structure in \cref{eq:reservoir} is fixed. Then, the reservoir is fed
an example input $\mathbf{u}(t)$ for which we know the desired output
$\mathbf{y_\text{desired}}(t)$. This example input produces a
reservoir response $\mathbf{r}(t)$ via \cref{eq:reservoir}. Then, we
choose $W_\text{out}$ to minimize the difference between
$\mathbf{y}(t)$ and $\mathbf{y_\text{desired}}(t)$, to approximate
\begin{equation}
  \label{eq:outputapprox}
  \mathbf{y_\text{desired}}(t) \approx W_\text{out} \mathbf{\tilde{r}}(t).
\end{equation}
More details of how this approximation is performed can be found in
\cref{subsec:train}.

Once the reservoir computer is trained, \cref{eq:reservoir,eq:output}
describe the complete process to transform the RC's
input $\mathbf{u}(t)$ into its output $\mathbf{y}(t)$.

\subsection{Forecasting}

To forecast a signal $\textbf{u}(t)$ with an RC, we construct the RC as usual, but train $W_\text{out}$
to reproduce the reservoir input $\textbf{u}(t)$: we set
$W_\text{out}$ to best approximate
\begin{equation}
  \label{eq:outputforecast}
  \mathbf{u}(t) \approx W_\text{out} \mathbf{\tilde{r}}(t).
\end{equation}
To begin forecasting, we replace the
input to the RC with the output. That is, we replace
$\mathbf{u}(t)$ with $W_\text{out}\mathbf{\tilde{r}}(t)$, and \cref{eq:reservoir} with
\begin{equation}
  \label{eq:reservoirforecast}
  \dot{\mathbf{r}}(t) = -\gamma\,\mathbf{r}(t) + \gamma\,\tanh \left( W_r \mathbf{r}(t) + W_\text{in} W_\text{out} \mathbf{\tilde{r}}(t) \right),
\end{equation}
which no longer has a dependence on the input
$\mathbf{u}(t)$ and runs autonomously. If $W_\text{out}$ is
chosen well, then $W_\text{out}\mathbf{\tilde{r}}(t)$
will approximate the original input $\mathbf{u}(t)$. These two signals
can be compared to assess the quality of the forecast (see \cref{sec:forecasting}).

\section{Example Systems}%
\label{sec:systems}

\hlchangeI{As examples for the forecasting task, we consider three
  chaotic systems: Lorenz '63, the R{\"{o}ssler} system, and a
  double-scroll chaotic circuit. Because all three of these systems are
  three-dimensional, they can all be used as inputs to the same
  reservoir computer without modifying $W_{\text{in}}$.}
  
\hlchangeI{To ensure that the results for all three systems are
  directly comparable, we rescale the temporal axis so that the
  maximum positive Lyapunov exponent matches that of the Lorenz
  system, $\lambda = 0.9056$. We also shift and rescale each component
  of the system to have zero mean and unit variance, to finally
  produce the true three-dimensional reservoir input $\mathbf{u}(t)$.}

\subsection{Lorenz '63}%

\begin{figure}
  \includegraphics{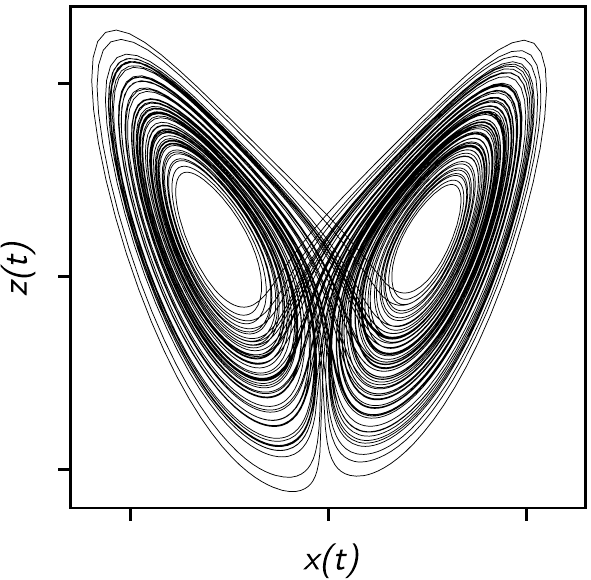}
  \caption{The true Lorenz attractor in the $x$/$z$ plane, produced by integrating \cref{eq:lorenz}.}%
  \label{fig:lorenz}
\end{figure}

The Lorenz '63 chaotic system is described by
\begin{equation}
  \begin{aligned}
    \dot{x} &= 10 \left(y - x\right), \\
    \dot{y} &= x \left(28 - z\right) - y, \\
    \dot{z} &= x\,y - \frac{8}{3} z,
  \end{aligned}
  \label{eq:lorenz}
\end{equation}
with standard parameters.\cite{lorenz1963} The attractor of this
system can be visualized easily in two dimensions by projecting the
three-dimensional trajectory of the system onto a plane. We show the
attractor in the $x$/$z$ plane in \cref{fig:lorenz}.

Our goal is to train an RC by training on a segment of the Lorenz
dynamics with \cref{eq:reservoir}, then perform prediction of the
Lorenz system after that segment with
\cref{eq:reservoirforecast}. Because the Lorenz system is chaotic,
forecasting must eventually fail. We choose to only perform prediction
for windows of one Lyapunov period $1 / \lambda = 1.104$ when evaluating
reservoir performance.

\subsection{R{\"{o}}ssler}%

\hlchangeI{The R{\"{o}}ssler system is described by
\begin{equation}
  \begin{aligned}
    \dot{x} &= - y - z, \\
    \dot{y} &= x + a y, \\
    \dot{z} &= b + z (x - c),
  \end{aligned}
  \label{eq:rossler}
\end{equation}
with standard parameters $a = 0.2$, $b = 0.2$, $c = 5.7$.\cite{rossler1976}}

\hlchangeIr{The $z$ component of this system mostly stays near zero,
  with rare positive spikes. This makes prediction with an RC
  difficult. To make this component of the system more suitable for RC
  prediction, we use $\log z$ instead for both RC input and prediction
  output.}\label{txt:logz}

\subsection{Double-Scroll}

\hlchangeI{The double-scroll chaotic circuit is described by the dimensionless equations
\begin{equation}
 \begin{aligned}
   \dot{V_1} &= \frac{V_1}{R_1} - \frac{V_1 - V_2}{R_2} - 2 I_r \sinh\left(\alpha(V_1 - V_2)\right), \\
   \dot{V_2} &= \frac{V_1 - V_2}{R_2} + 2 I_r \sinh\left(\alpha(V_1 - V_2)\right) - I, \\
   \dot{I} &= V_2 - R_4 I,
 \end{aligned}
 \label{eq:dscroll}
\end{equation}
with parameters $R_1 = 1.2$, $R_2 = 3.44$, $R_4 = 0.193$, $I_r = 2.25
\times 10^{-5}$, and $\alpha = 11.6$.\cite{gauthier1996}}

\section{Reservoir Construction and Training}%
\label{sec:construct-train}

To build our reservoir computers, we need to build the internal
network to use as the reservoir, create connections from the nodes to
the overall input, and then train it to fix $W_\text{out}$. Once this
is completed, the RC will be fully specified and able to perform
forecasting.

\subsection{Internal Reservoir Construction}%
\label{subsec:construct}

There are many possible choices for generating the internal reservoir
connections $W_r$ and the input connections $W_\text{in}$. For
$W_\text{in}$, we randomly connect each node to each
RC input with probability $\sigma$. The weight for
each connection is drawn randomly from a normal distribution with mean
$0$ and variance $\rho_\text{in}^2$. Together, $\sigma$ and
$\rho_\text{in}$ are enough to generate a random instantiation of
$W_\text{in}$.

\begin{figure*}
  \includegraphics[width=\textwidth]{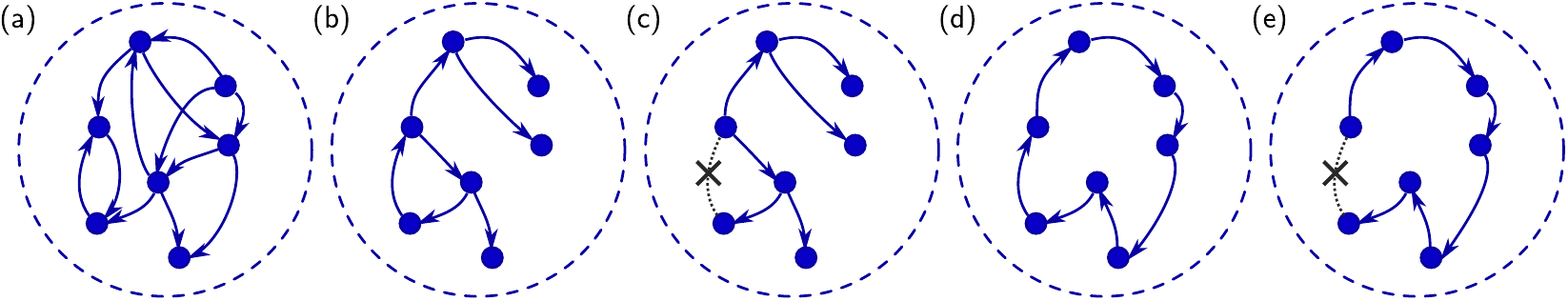}
  \caption{The five reservoir topologies tested. Only internal
    reservoir connections are pictured. Connections to the reservoir
    computer input, or to the output layer (as in \cref{fig:overview})
    are not shown. (a) A general, fixed in-degree network, here
    pictured with $N=7$ and $k=2$. (b) A $k=1$ network with a single
    connected component. (c) A $k=1$ network with the single cycle cut
    at an arbitrary point. (d) A \emph{simple cycle reservoir}. (e) A
    \emph{delay line reservoir}.}%
  \label{fig:topology}
\end{figure*}

For the internal connections $W_r$, we generate a random network where
every node has a fixed in-degree $k$. For each node, we select $k$
nodes in the network without replacement and use those as inputs to
the current node. Each input is assigned a random weight drawn from a
normal distribution with mean $0$ and variance $1$. This results in a
connection matrix $W_r'$ where each row has exactly $k$ non-zero
entries. Finally, we rescale the whole matrix
\begin{equation}
  \label{eq:setradius}
  W_r = \frac{\rho_r}{\operatorname{SR}(W_r')} W_r',
\end{equation}
where $\operatorname{SR}(W_r')$ is the spectral radius, or maximum
absolute eigenvalue, of the matrix $W_r'$. This scaling ensures that
$\operatorname{SR}(W_r) = \rho_r$. Together, $k$ and $\rho_r$ are
enough to generate a random instantiation of $W_r$. We present an
example of such a network in \cref{fig:topology}~(a).

Therefore, to create a random instantiation of a RC
suitable to begin the training process, we must set a value for five
\emph{hyperparameters}:

\begin{itemize}
\item $\gamma$, which sets the characteristic time scale of the reservoir,
\item $\sigma$, which determines the probability a node is connected to a reservoir input,
\item $\rho_\text{in}$, which sets the scale of input weights,
\item $k$, the recurrent in-degree of the reservoir network,
\item $\rho_r$, the spectral radius of the reservoir network.
\end{itemize}
We select these parameters by searching a range of acceptable values selected to minimize the forecasting error using the Bayesian optimization procedure. The
details of this can be found in \cref{subsec:bayes}. However, during
the optimization process, we discovered that the optimizer was often
finding RCs with $k = 1$ that perform as well as
RCs with a higher $k$. Such reservoirs have an
interesting and simple network topology, thereby suggesting other
simple topologies for comparison.

First, networks with $k = 1$ generated with our algorithm often
have disconnected components. These components essentially act as
RCs operating in parallel; we do not consider these further even though it is an interesting line of research.\cite{pathak2018} We
limit ourselves to only looking at reservoir networks with a single
connected component.

If a $k = 1$ network only has a single connected component, then it
must also contain only a single directed cycle. This limits how
recurrence can occur inside the network compared to higher-$k$
networks. Every node in a $k = 1$ network is either part of this cycle
or part of a directed tree branching off from this cycle, as depicted
in \cref{fig:topology}~(b). Inspired by the high performance of this
simple structure, we also investigate $k = 1$ networks when the single
cycle is cut at an arbitrary point. This turns the entire network into
a tree, as in \cref{fig:topology}~(c).

Finally, we also investigate reservoir networks that consist entirely
of a cycle or ring with identical weights with no attached tree
structure, depicted in \cref{fig:topology}~(d), and networks with a
single line of nodes (a cycle that has been cut), in
\cref{fig:topology}~(e). These are also known as \emph{simple cycle
  reservoirs} and \emph{delay line reservoirs},
respectively.\cite{rodan2011}

In total, there are five topologies we investigate:
\begin{enumerate}[label= (\alph*)]
\item general construction with unrestrained $k$,
\item $k = 1$ with a single cycle,
\item $k = 1$ with a cut cycle,
\item single cycle, or \emph{simple cycle reservoir},
\item single line, or \emph{delay line reservoir}.
\end{enumerate}
Both the $k = 1$ cut cycle networks (c) and line networks (e) are
rescaled to have a fixed $\rho_r$ before the cycle is
cut. However, after the cycle is cut, they both have $\rho_r=0$.

\subsection{Bayesian Optimization}%
\label{subsec:bayes}

The choice of hyperparameters that best fits this problem is
difficult to identify. Grid search\cite{rodan2011} and gradient descent\cite{jaeger2007} have been used previously. However, these algorithms
struggle with either non-continuous parameters or noisy results. Because
$W_r$ and $W_\text{in}$ are determined randomly, our optimization
algorithm should be able to handle noise.  We use Bayesian optimization,\cite{yperman2016,maat2018} as
implemented by the \texttt{skopt} Python
package.\cite{skopt2018} Bayesian optimization deals well with both
noise and integer parameters like $k$, is more efficient than grid
search,\cite{maat2018} and works well with minimal tuning.

\begin{table}
  \caption{Range of hyperparameters searched using Bayesian optimization.}
  \begin{tabular}{lrcl}
    Parameter & min & & max \\
    \hline
    $\gamma$ & 7 & -- & 11 \\
    $\sigma$ & 0.1 & -- & 1.0 \\
    $\rho_\text{in}$ & 0.3 & -- & 1.5 \\
    $k$ & 1 & -- & 5 \\
    $\rho_r$ & 0.3 & -- & 1.5 \\
  \end{tabular}%
  \label{tab:ranges}
\end{table}
  
For each topology, the Bayesian algorithm repeatedly generates a set
of hyperparameters to test within the ranges listed in
\cref{tab:ranges}. Larger ranges require a longer optimization
time. We selected these ranges to include the values that existing
heuristics would choose, and to allow exploration of the space without
a prohibitively long runtime.  However, exploring outside these ranges
is valuable. Here we focus on the connectivity $k$, but expanding the
search range for the other parameters may also produce interesting
results.

At each iteration of the algorithm, the optimizer constructs a single
random reservoir with the chosen hyperparameters, trains it according
to the procedure described in \cref{subsec:train}, and measures its
performance with the metric $\epsilon$ described in
\cref{sec:forecasting}. From this measurement, it chooses a new set of
hyperparameters to test that may be closer to the optimal values. We
limit the number of iterations of this algorithm to test a maximum of
100 reservoir realizations before returning an optimized reservoir. In
order to estimate the variance in the performance of reservoirs
optimized by this method, we repeat this process 20 times.

\subsection{Training}%
\label{subsec:train}

To train the RC, we integrate
\hlchangeI{\cref{eq:reservoir} coupled with the chosen input
  (\cref{eq:lorenz,eq:rossler,eq:dscroll})}
via the hybrid Runge-Kutta 5(4)\cite{dormand1980}
method from $t = 0$ to $300$ with a fixed time step
$\Delta t = 0.01$, and divide this interval into three ranges:
\begin{itemize}
\item $t = 0$ -- $100$: a transient, which is discarded,
\item $t = 100$ -- $200$: the training period,
\item $t = 200$ -- $300$: the testing period.
\end{itemize}
We use the transient period to ensure the later times are not
dependent on the specific initial conditions. We divide the rest into
a training period, used only during training, and a testing period,
used later only to evaluate the RC performance. 

This integration produces a solution for $\mathbf{r}(t)$. However,
\hlchangeI{when the reservoir is combined with the Lorenz system, it has}
a symmetry that can confuse
prediction.\cite{pathak2017} Before integration, we break this symmetry by setting $\mathbf{f_\text{out}}$ so that
\begin{equation}
  \tilde{r}_i(t) = \begin{cases}
    r_i(t) & \text{if } i \leq N / 2, \\
    r_i{(t)}^2 & \text{if } i > N / 2.
  \end{cases}
  \label{eq:reservoirsym}
\end{equation}
\hlchangeI{For consistency across our three example input systems, this is done
  for every reservoir we construct, even if the input system we
  eventually use is not the Lorenz system.}
We then find a $W_\text{out}$ to minimize
\begin{equation}
  \sum_{t=100}^{200} |\mathbf{u}(t) - W_\text{out}\,\mathbf{\tilde{r}}(t)|^2 + \alpha ||W_\text{out}||^2 ,
  \label{eq:ridge}
\end{equation}
where the sum is understood to be over time steps $\Delta t$ apart.  Now that $W_\text{out}$ is determined, the RC is trained.

Equation~\ref{eq:ridge} is known as Tikhonov regularization, or ridge regression. The
ridge parameter $\alpha$ could be included among the hyperparameters
to optimize. However, unlike the other hyperparameters, modifying
$\alpha$ does not require re-integration and can be optimized with
simpler methods. We select an $\alpha$ from among $10^{-5}$ to $10^5$
by leave-one-out cross-validation. This also reduces the number of
dimensions the Bayesian algorithm must work with.

\section{Forecasting and Evaluation}%
\label{sec:forecasting}

To evaluate the performance of the trained RC, we use it to perform autonomous
forecasting using \cref{eq:reservoirforecast}.

The most common method for evaluating an RC forecast is
to choose a time $t_1$ to begin forecasting and then compare
the forecast to the true system.\cite{lukosevicius2012} Usually, $t_1$ is chosen to be directly after the training period, which is $t_1 =200$
in our procedure. Then, we initialize the reservoir state $\mathbf{r}(t_1)$ to the
value found earlier during training and integrate
\cref{eq:reservoirforecast} for one Lyapunov time, between $t =
t_1$ and $t = t_1 + 1/\lambda$. This produces a reservoir forecast
$W_\text{out}\;\mathbf{\tilde{r}}(t)$ during these times, which we
compare to the true system $\mathbf{u}(t)$ to produce a
root-mean-squared error (RMSE)
\begin{equation}
  \epsilon_1 = {\left( \Delta t\,\lambda \sum_{t=t_1}^{t_1 + 1/\lambda} |\mathbf{u}(t) - W_\text{out} \mathbf{\tilde{r}}(t)|^2 \right)}^{1/2}.
  \label{eq:rmse}
\end{equation}
Note that $\epsilon_1$ is normalized because
we construct the input signal
$\mathbf{u}(t)$ with unit variance.

\begin{figure}
  \includegraphics{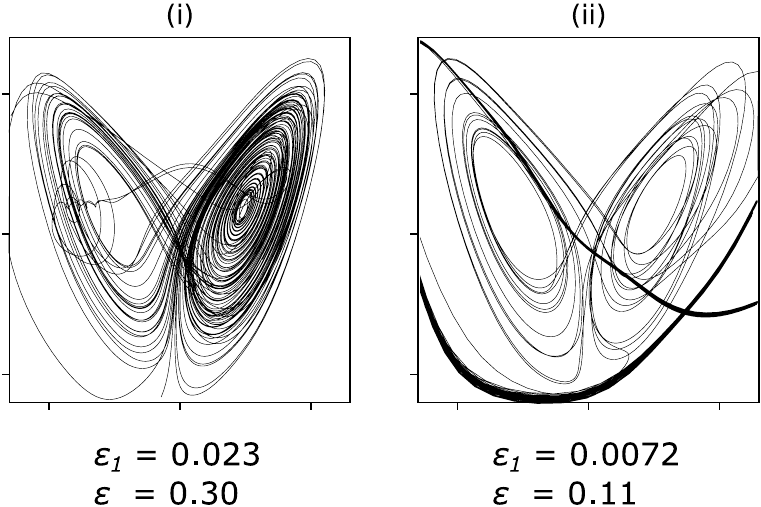}
  \caption{Two examples of reservoir computers that fail to reproduce
    the Lorenz climate, but produce a low $\epsilon_1$. Compare with
    the true attractor in \cref{fig:lorenz}. Reservoir (i) fails to
    learn the attractor, overemphasizing the right lobe in which $t_1$
    resides. Reservoir (ii) matches the attractor well early on, but
    as the prediction lengthens, it falls into a periodic orbit (thick
    black line). Both (i) and (ii) show a promising low $\epsilon_1$,
    but the averaged $\epsilon$ measure more accurately captures their
    failure to learn the Lorenz attractor.}%
  \label{fig:failure}
\end{figure}

This method of evaluating forecasting ability is flawed for our
purposes. Previous results have shown that a low $\epsilon_1$ is not a
good indicator of whether a reservoir computer has learned the climate
of a system\cite{pathak2017,haluszczynski2019} and we also observe the same effect here.
\Cref{fig:failure} depicts two common ways for an RC to fail to
replicate the true Lorenz attractor, shown in \cref{fig:lorenz}.  However,
both produce a good short-term forecast near $t_1$. In particular,
reservoir (ii) in \cref{fig:failure} has a lower $\epsilon_1$ score
than any of the optimized reservoirs despite its obvious
failure to learn the Lorenz attractor.

This problem is exacerbated in an optimization setting because 
searching for reservoirs with the lowest $\epsilon_1$ risks producing
reservoirs that are only good at forecasts near $t_1$, but otherwise
perform poorly in reproducing the climate. It can also waste time, as the optimization algorithm explores areas of the parameter space it believes perform well, but
actually do not.

Finally, the choice of $t_1$ can dramatically affect $\epsilon_1$. For example, the
Lorenz system has an unstable saddle point at the origin, and
trajectories that approach this point may end up in either the left or
right lobe of the attractor in \cref{fig:lorenz}. If $t_1$ happens to
lie near this point, then even very small prediction errors can be
magnified. For example, if the reservoir predicts a trajectory into
the left lobe, while the true system goes to the right, the
$\epsilon_1$ measure might be very high even though the reservoir is
well-trained.

To combat these problems, we instead evaluate a short-term forecast at
$50$ times $t_i$, evenly spaced within our testing period between
$t=200\;\text{--}\;300$. For each $t_i$, we perform the evaluation method as
described above, producing $50$ error measures
$\epsilon_i$. Because each $t_i$ is drawn from the testing period, we
only evaluate the reservoir on data it has not yet seen: no information
about the input or reservoir system at $t_i$ is used to construct
$W_\text{out}$.

We then combine these errors into a single overall error
\begin{equation}
  \epsilon = {\left( \frac{1}{50} \sum_{i = 1}^{50} \epsilon_i^2 \right)}^{1/2}
\end{equation}
that represents the average ability of the reservoir computer to
forecast accurately at any point on the input system attractor.

\begin{figure}
  \includegraphics{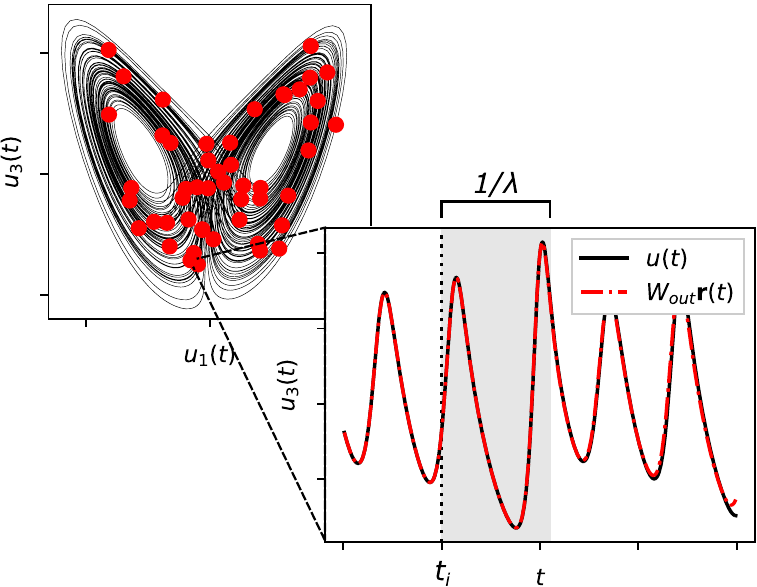}
  \caption{Summary of our forecasting evaluation method. We calculate
    errors $\epsilon_i$ at times $t_i$, marked on the attractor as red
    dots, and combine them into a single error $\epsilon$. Before each
    $t_i$ (dotted vertical line), the reservoir is integrated with
    \cref{eq:reservoir} and listening to the input. After
    $t_i$, the reservoir is integrated with
    \cref{eq:reservoirforecast}, and runs autonomously. The
    reservoir's prediction (dotted red line) must eventually diverge
    from the true system (solid black line). We calculate $\epsilon_i$
    only during a single Lyapunov period after forecasting begins,
    marked here in grey.}%
  \label{fig:testing}
\end{figure}

The parameter $\epsilon$ is our figure of merit that the Bayesian algorithm is
trying to minimize. By combining short-term forecasting errors from
many points on the input attractor, this metric emphasizes learning
the global dynamics of the system as opposed to short-term
forecasting over a single temporal segment.
\hlchangeII{Using this method, we see that the Bayesian optimization
  algorithm works more effectively, as it no longer gets trapped in
  valleys of low $\epsilon_1$ that otherwise fail to reproduce the
  attractor.}\label{txt:optimize-better}
This evaluation method is summarized in \cref{fig:testing}.

\section{Results}%
\label{sec:results}

We run all five reservoir topologies through 100 iterations of the
Bayesian algorithm
\hlchangeI{using the Lorenz system as input,}
and record the best-performance RC for each
topology according to the metric $\epsilon$. These reservoirs, and the
hyperparameters that generated them, are reported in
\cref{tab:results}. We estimate the errors on these with the standard
deviation of $\epsilon$ after repeating the optimization process 20
times.

\begin{table}
  \caption{Best reservoir computers of each topology, after 100
    iterations of the Bayesian optimization algorithm
    \hlchangeI{using the Lorenz system as input.}
    The
    hyperparameters chosen by the algorithm are shown on the
    right. The simpler topologies (b) -- (e) all perform nearly as
    well as the general topology (a).}
  \begin{ruledtabular}
    \begin{tabular*}{\linewidth}{l l@{\extracolsep{\fill}} l d d d d}
      & & \multicolumn{5}{l}{Lorenz} \\
      & Topology & $\epsilon$ & \multicolumn{1}{c}{$\gamma$} & \multicolumn{1}{c}{$\sigma$} & \multicolumn{1}{c}{$\rho_\text{in}$} & \multicolumn{1}{c}{$\rho_r$} \\
      \hline
      (a) & Any $k$ \footnote{After optimization, $k = 3$.} & 0.022 $\pm$ 0.004 & 7.7 & 0.81 & 0.37 & 0.41 \\
      (b) & $k = 1$ with cycle & 0.024 $\pm$ 0.005 & 10.9 & 0.44 & 0.30 & 0.30 \\
      (c) & $k = 1$ no cycle & 0.028 $\pm$ 0.005 & 7.2 & 0.78 & 0.30 & 0.30 \footnote{$\rho_r$ measured before cycle is cut. Afterwards, $\rho_r = 0.$} \\
      (d) & cycle & 0.023 $\pm$ 0.008 & 7.9 & 0.17 & 0.58 & 0.30 \\
      (e) & line & 0.024 $\pm$ 0.003 & 10.6 & 0.79 & 0.30 & 0.45 \footnotemark[2] \\
    \end{tabular*}
  \end{ruledtabular}%
  \label{tab:results}
\end{table}

When optimized, all reservoir topologies perform well. In particular,
the simpler topologies all perform almost as well as the general-$k$
topology. They often lie within one, or at most two standard
deviations from topology (a). This is despite the fact that topologies
(c) and (e) both have $\rho_r=0$ and no recurrent connections within
the network. The other topologies have $\rho_r\ll1$.
\hlchangeII{Previous work has already demonstrated that reservoirs with
  low or zero spectral radius can still
  function.\cite{pathak2017,rodan2011} These results act as additional
  counterexamples to the heuristic that reservoir computers should
  have $\rho_r \approx 1$.\cite{lukosevicius2012}}%
\label{txt:heuristics1}

However, these best-observed reservoirs are not representative of a
typical RC.\@ We use the hyperparameters to guide the random input
connections and connections within the reservoir, and so even once the
hyperparameters are fixed, constructing the reservoir is a random
process. Not all reservoirs with a fixed topology and hyperparameters
will perform the same.

To explore this variation, we generate and evaluate $200$ RCs of each
topology \hlchangeI{on the Lorenz system,} using the optimized
hyperparameters in \cref{tab:results}.  \hlchangeII{For all five
  topologies, as the measured $\epsilon$ increases, the quality of the
  reproduced attractor decreases gradually.  \hlchangeIIr{On manual
    inspection of the attractors, the reason for this decrease can be
    divided into three qualitative regions. For $\epsilon < 0.3$, the
    RCs reproduce the Lorenz attractor consistently. Failures still
    rarely occur, but they always reproduce part of the attractor
    before falling into a fixed point or periodic orbit. In this
    region, small differences between the true attractor and the
    reproduced attractor contribute more to $\epsilon$ than outright
    failure. For $0.3 < \epsilon < 1.0$, RCs always fail to reproduce
    the attractor, though they will still always reproduce a portion
    of it before failing. Examples of these failures are provided in
    \cref{fig:failure}. Above $\epsilon > 1.0$, these failures become
    catastrophic, and no longer resemble the Lorenz attractor at
    all. A more quantitative description of these regions is one line
    of possible future research.\label{txt:examples-prediction}}}

\begin{figure}
  \includegraphics{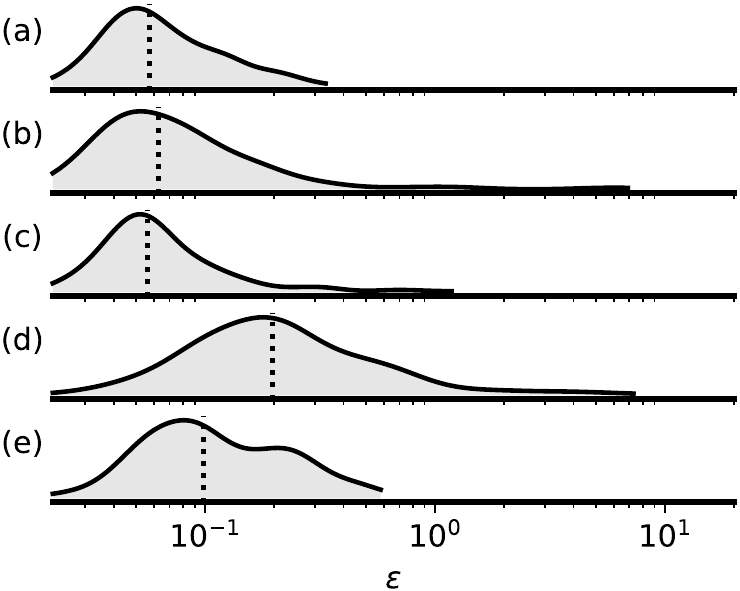}
  \caption{Performance for each reservoir topology, evaluated
    \hlchangeI{for the Lorenz system}
    at the
    hyperparameters listed in \cref{tab:results}.
    \hlchangeII{Each is visualized as a Gaussian kernel density
      estimation in $\log_{10} \epsilon$ with a bandwidth of
      0.35,\cite{scott1992} which can be interpreted as a probability
      distribution. Using a narrower bandwidth does not reveal additional
      features.}
    A vertical line marks the median.  Note that topologies
    (b) -- (e) have very long tails, and can produce reservoirs that
    perform very poorly in comparison to (a). However, all five have
    the capability to produce well-performing reservoirs.}%
  \label{fig:distribution}
\end{figure}

Though the optimized best-performing reservoirs of
each topology show very little performance difference, the differences
between them become more apparent when we compare the probability
distribution of $\epsilon$ for each topology. These
distributions are shown in \cref{fig:distribution}.

A well-performing reservoir with arbitrary $k$ (a) is a much more
likely outcome than a well-performing reservoir with a single cycle
(d). However, the performance of arbitrary $k$ reservoirs (a) is very
similar to that of tree-like reservoirs (c). Though (c) has a longer
tail on the high end, the simpler structure of the reservoir may be
appealing for hardware RCs.

The distribution of performance for each topology can be a deciding
factor if reservoir construction and evaluation is expensive, as it
might be in hardware. In software, though, we find the
best-performing reservoirs in \cref{tab:results} after only $100$
trials. A hardware design can still benefit from the simpler
topologies (b) -- (e) despite their very wide performance distributions
if the creation of the evaluation of the design can be automated to
test many candidate reservoirs, as on an FPGA.\cite{canaday2018}

\begin{figure*}
  \includegraphics[width=\textwidth]{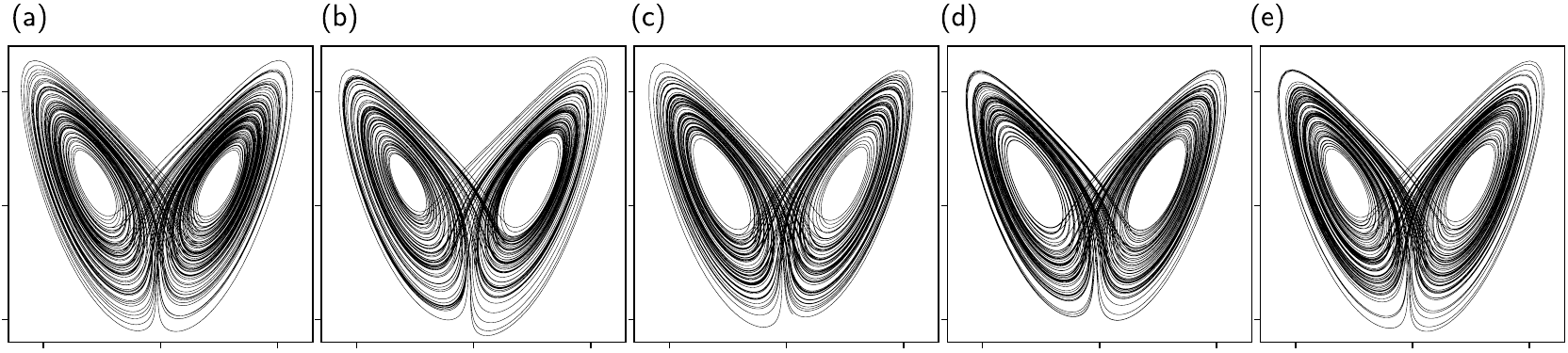}
  \caption{Lorenz attractor plots in the $x$/$z$ plane for long-term
    free-running predictions for each reservoir topology in \cref{tab:results}. Compare with
    true Lorenz attractor in \cref{fig:lorenz}, and the failed
    reservoirs in \cref{fig:failure}.}%
  \label{fig:attractors}
\end{figure*}

There may also be a benefit in software. The simpler topologies are
represented by weight matrices $W_r$ in very simple forms. Topology
(c) can always be represented as a strictly lower-diagonal matrix, and
(d) -- (e) can be represented with non-zero entries only directly below
the main diagonal. Software simulations can take advantage of this structure to
speed up integration of \cref{eq:reservoir}.

\hlchangeI{To explore whether these topologies remain equally
  effective at tasks beyond forecasting the Lorenz system, we run
  all five reservoir topologies through 100 iterations of the Bayesian
  algorithm for both the R{\"{o}}ssler and the double-scroll
  systems. As with the Lorenz system, we estimate the errors on these
  results by repeating the process 20 times each. These results are
  reported in \cref{tab:resultsplus}.}

\begin{table}
  \caption{\hlchangeI{Result of optimizing reservoirs with the
      Bayesian algorithm over 100 iterations, on both the
      R{\"{o}}ssler system and the double-scroll system. All five
      topologies can perform equally well at both systems.}}
  \begin{ruledtabular}
    \begin{tabular*}{\linewidth}{l l@{\extracolsep{\fill}} l l}
      & & \multicolumn{1}{l}{Double Scroll} & \multicolumn{1}{l}{R{\"{o}}ssler} \\
      & Topology & $\epsilon$ & $\epsilon$ \\
      \hline
      (a) & Any $k$ & 0.029 $\pm$ 0.006 & 0.017 $\pm$ 0.005 \\
      (b) & $k = 1$ with cycle & 0.033 $\pm$ 0.007 & 0.020 $\pm$ 0.007 \\
      (c) & $k = 1$ no cycle & 0.033 $\pm$ 0.008 & 0.018 $\pm$ 0.006 \\
      (d) & cycle & 0.033 $\pm$ 0.007 & 0.018 $\pm$ 0.006 \\
      (e) & line & 0.037 $\pm$ 0.01 & 0.019 $\pm$ 0.015
    \end{tabular*}
  \end{ruledtabular}%
  \label{tab:resultsplus}
\end{table}

\hlchangeI{The results for the double-scroll \hlchangeIr{and
    R{\"{o}}ssler systems} agree well with those for Lorenz. All five
  topologies optimize reliably with the Bayesian algorithm, and all
  perform similarly when optimized. Optimizing a reservoir to
  reproduce \hlchangeIr{either system} will almost always work after
  only 100 iterations.}

\label{txt:newresults}

\hlchangeI{One advantage to RCs is that a single reservoir can be
  re-used on many different tasks by re-training $W_{\text{out}}$. To
  evaluate whether this is possible with these optimized reservoirs,
  we take the 20 reservoirs optimized for the Lorenz system and
  re-train $W_{\text{out}}$ for each to instead reproduce the
  double-scroll circuit system. Every other part of the RC is left
  unchanged. We then evaluate how accurate this prediction is using
  the metric $\epsilon$. These results are summarized in
  \cref{tab:resultsgen}.}

\begin{table}
  \caption{\hlchangeI{Result of re-using reservoirs optimized for
      Lorenz prediction to perform double-scroll prediction. The
      minimum error encountered across all 20 reservoirs of each
      topology is reported as $\epsilon_{\text{min}}$.}}
  \begin{ruledtabular}
    \begin{tabular*}{\linewidth}{l l@{\extracolsep{\fill}} l l}
      & & \multicolumn{2}{l}{Double Scroll} \\
      & Topology & $\epsilon$ & $\epsilon_\text{min}$ \\
      \hline
      (a) & Any $k$ & 0.43 $\pm$ 1.2 & 0.028\\
      (b) & $k = 1$ with cycle & 0.30 $\pm$ 0.5 & 0.048 \\
      (c) & $k = 1$ no cycle & 0.37 $\pm$ 0.8 & 0.032 \\
      (d) & cycle & 0.17 $\pm$ 0.2 & 0.056 \\
      (e) & line & 0.22 $\pm$ 0.3 & 0.058
    \end{tabular*}
  \end{ruledtabular}%
  \label{tab:resultsgen}
\end{table}

\hlchangeI{In general, these reservoirs perform poorly on this new
  task. However, there is extremely high variation. Even though they
  were optimized to perform Lorenz forecasting, many of these
  reservoirs are still able to reproduce the double-scroll
  attractor. Moreover, the best performers in each category approach
  the performance of reservoirs optimized specifically for the
  double-scroll system. This indicates that it is possible to find a
  single reservoir in any of these topologies that works well on more
  than one system. The Bayesian optimization algorithm may be able to
  find these reservoirs more reliably if the metric $\epsilon$ is
  modified to reward RCs that perform well on many systems.}

Finally, for each topology produced a free-running prediction
\hlchangeI{of the Lorenz system}
for $100$ time units using the best-performing RC.\@ We use these predictions to produce an attractor as shown in \cref{fig:attractors}. All five optimized
RCs reproduce the Lorenz attractor well. Though
comparing these plots by eye is not quantitative, we find them
qualitatively useful: an RC that fails to reproduce the
Lorenz attractor by eye is unlikely to have a low $\epsilon$ compared
to one that does.

\section{Conclusion}%
\label{sec:conclusion}

We find that Bayesian optimization of RC
hyperparameters is a useful tool for creating high-performance reservoirs
quickly. We also find that allowing the optimizer to explore
areas of the parameter space that are typically excluded in other optimization studies can lead to interesting and effective reservoir designs such as those presented
here. 

For this procedure to be effective, we find that evaluating the RC performance at many points along the attractor and averaging, rather than at a single point,
encourages the optimization algorithm to find reservoirs that
reproduce the Lorenz climate. Using this evaluation method helps
direct the optimizer away from reservoirs that perform good short-term
forecasting at only one point on the Lorenz attractor.

One surprising outcome of our optimization procedure is finding reservoirs that perform well even with no recurrent
connections and $\rho_r=0$.
\hlchangeII{Though some reservoirs of this kind have been explored previously
  and shown to work,\cite{pathak2017,rodan2011} the heuristics remain
  common in reservoir design. We present additional concrete examples
  that provide evidence these heuristics are not unbreakable rules.}%
\label{txt:heuristics2}

In greater detail, we find reservoirs with very low internal
connectivity that perform at least as well as their
higher-connectivity counterparts. A reservoir with only a single
internal cycle, or even no cycle at all, can perform as well as those
with many recurrent cycles. These simpler topologies manifest as
simpler weight matrices, which can result in faster integration in
software. In a hardware environment where connections between nodes
have a cost, or recurrence is difficult to implement, these network
topologies may also have a direct benefit.

Though the best of these low-connectivity reservoirs perform as well
as the more complicated reservoirs, they tend to perform worse on
average. However, searching for the best-performing instance of these
reservoirs can be done in few trials, and may be feasible for hardware
reservoirs that can be constructed and evaluated in an automated way.

\hlchangeII{We have discovered many interesting lines of future
  research. First, we can evaluate the new metric $\epsilon$ by
  comparing the output of the reservoir predictions to the true system
  attractor using a new metric for attractor
  overlap.\cite{ishar2019}}
  \hlchangeIIr{This overlap metric can also be used to quantify the qualitative observations of different failure modes in regions of our $\epsilon$ metric.\label{txt:regions-metric}}
  \hlchangeI{Second, there are known results
  that prove that a linear network architecture with time-independent nodes, the
  discrete-time NARX networks, can simulate fully-connected networks.\cite{siegelmann1997}
  There may
  be a similar proof for RCs, which might explain why
  we see no difference in the best-performing reservoirs in each
  topology. Third, our optimization procedure finds the best network
  weights $W_r$ for a given task. In many ways, this is a similar task
  to training a traditional recurrent neural network. We are
  interested in comparing this method to those used for networks other
  than RCs.}\label{txt:new-research}

Our results show that these very low connectivity reservoirs perform
well in the narrow context of software-based, chaotic system
forecasting. Future work will explore whether these results hold for
other reservoir computing tasks such as classification,
\hlchangeI{and whether it is possible to find reservoirs that perform well on a variety of tasks simultaneously by modifying the metric $\epsilon$ to encourage generalization.}\label{txt:generalization}
We also intend to
explore whether these results hold in hardware reservoirs and if the simpler reservoir designs allow for more
efficient and faster operating hardware RCs.

\begin{acknowledgments}
\hlchangeIr{We thank our reviewers for their comments, and for their
  suggestion to use $\log z$ for prediction in the R{\"{o}}ssler
  system.}\label{txt:logz-ack}
We gratefully acknowledge the financial support of the U.S.\ Army Research Office Grant No.\ W911NF-12-1-0099, DARPA Award No.\ HR00111890044, and a Network Science seed grant from the The Ohio State University College of Arts \& Sciences.
\end{acknowledgments}


\bibliography{paper}

\end{document}